\pdfoutput=1

\documentclass[11pt]{article}
\usepackage[dvipsnames,svgnames,table]{xcolor}

\usepackage[preprint]{acl}

\usepackage{times}
\usepackage{latexsym}
\usepackage{CJKutf8}
\usepackage{tabularx}
\usepackage{booktabs}
\usepackage{babel}
\usepackage{listings}
\usepackage[size=tiny, disable]{todonotes}

\usepackage[T1]{fontenc}

\usepackage[utf8]{inputenc}

\usepackage{microtype}

\usepackage{inconsolata}

\usepackage{graphicx}
\usepackage{amsmath}
\usepackage{amsfonts}
\usepackage{amssymb}

\newcommand{\ourmodel}[0]{\textsc{PromptOptMe}}
\newcommand{\se}[1]{\textcolor{black}{#1}}

\title{PromptOptMe: Error-Aware Prompt Compression for LLM-based MT Evaluation Metrics}

\author{
    Daniil Larionov\textsuperscript{1,2} \quad
    Steffen Eger\textsuperscript{1,3}\\
    \textsuperscript{1} NLLG, \textsuperscript{2} University of Mannheim, \textsuperscript{3} University of Technology Nuremberg\\
\href{mailto:daniil.larionov@uni-mannheim.de}{daniil.larionov@uni-mannheim.de}
}

\begin{document}
\maketitle
\setlength{\belowdisplayskip}{2pt} \setlength{\belowdisplayshortskip}{2pt}
\setlength{\abovedisplayskip}{2pt} \setlength{\abovedisplayshortskip}{2pt}
\begin{abstract}
Evaluating the quality of machine-generated natural language content is a challenging task in Natural Language Processing (NLP). Recently, large language models (LLMs) like GPT-4 have been employed for this purpose, but they are computationally expensive due to the extensive token usage required by complex evaluation prompts. In this paper, we propose a prompt optimization approach that uses a smaller, fine-tuned language model to compress input data for evaluation prompt, thus reducing token usage and computational cost when using larger LLMs for downstream evaluation. Our method involves a two-stage fine-tuning process: supervised fine-tuning followed by preference optimization to refine the model's outputs based on human preferences. We focus on Machine Translation (MT) evaluation and utilize the GEMBA-MQM metric as a starting point. Our results show a $2.37\times$ reduction in token usage without any loss in evaluation quality. This work makes state-of-the-art LLM-based metrics like GEMBA-MQM more cost-effective and efficient, enhancing their accessibility for broader use.
\end{abstract}

\section{Introduction}
\label{sec:introduction}

The rapid advancement of Natural Language Generation (NLG) technologies has led to an increasing reliance on automated systems for producing human-like text across various domains, including machine translation (MT). As these systems become more prevalent, the need for effective and efficient evaluation metrics to assess generated content quality has also grown. Evaluating NLG systems is a fundamental challenge in the field of Natural Language Processing (NLP).

\begin{figure}[h]
    \centering
    \includegraphics[width=\columnwidth]{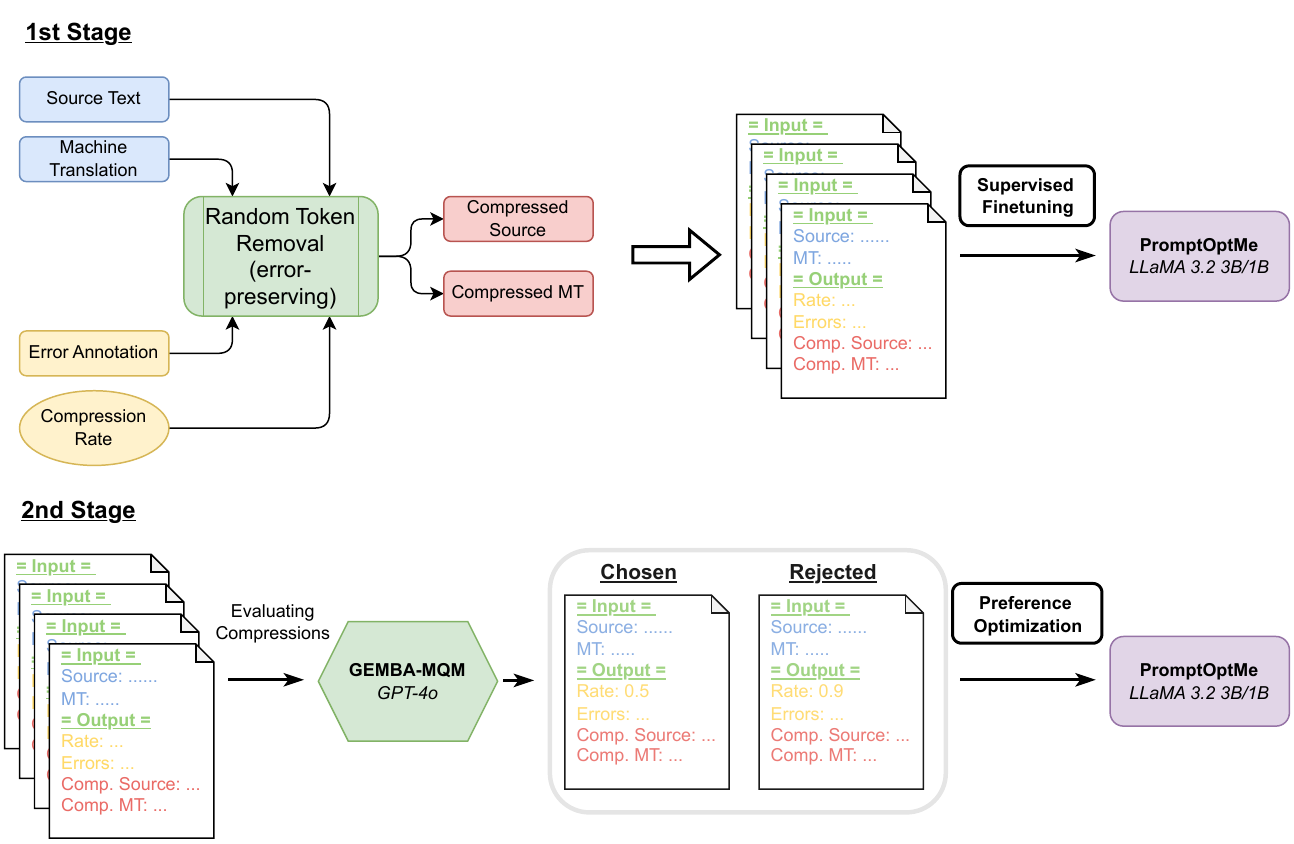}
    \caption{The two-stage model training approach used in PromptOptMe. At the first stage, the model is fine-tuned in a supervised way to adapt it for the compression task and prompt format. At the second stage, we utilize preference data obtained through evaluation of compressed prompts to train the model to select the best compression for each example.}
    \label{fig:method}
\end{figure}

Traditionally, automatic evaluation metrics such as BLEU~\citep{papineni-etal-2002-bleu}, ROUGE~\citep{lin-2004-rouge}, BERTScore~\citep{zhang2019bertscore}, and MoverScore~\citep{zhao-etal-2019-moverscore} have been widely used due to their simplicity and ease of implementation. BLEU and ROUGE measure n-gram overlap between the generated text and reference texts, but they often fail to capture the semantic meaning and penalize legitimate lexical variations~\citep{callison-burch-etal-2006-evaluating}. BERTScore and MoverScore leverage contextual embeddings from pre-trained language models to compute similarity at a deeper semantic level; however, they can still struggle with capturing nuanced errors in meaning and may not effectively evaluate aspects like factual correctness or language fluency~\citep{freitag-etal-2021-experts, kocmi-etal-2021-ship, chen-eger-2023-menli}.

Subsequently, trained evaluation metrics such as COMET~\citep{rei-etal-2022-comet}, BLEURT~\cite{sellam-etal-2020-bleurt}, and xCOMET~\citep{guerreiro2023xcomet} emerged. These models are trained on human-annotated datasets to predict quality scores, allowing them to better align with human judgments. Despite their improved quality, these metrics require substantial amounts of labeled data for training and may not generalize well across different tasks or domains.

With the advent of large language models (LLMs) like GPT-4~\citep{achiam2023gpt}, researchers have started to explore their potential for NLG evaluation through prompting instead of traditional training. Metrics such as GEMBA-DA~\citep{kocmi-federmann-2023-large}, GEMBA-MQM~\cite{kocmi-federmann-2023-gemba}, AutoMQM~\citep{fernandes-etal-2023-devil}, and G-Eval~\citep{liu-etal-2023-g} utilize LLMs by providing instructions and few-shot examples within prompts. This approach offers several key benefits: {\bf a)} elimination of task-specific training, as LLMs can perform evaluations based on prompts without the need for labeled datasets; {\bf b)} high-quality results, with LLM-based metrics demonstrating superior alignment with human judgments and more effectively capturing linguistic nuances and contextual information; {\bf c)} flexibility and generalization, since these models can be easily adapted to different tasks and domains by modifying prompts, offering greater flexibility compared to traditional trained metrics.

Despite these advantages, using LLMs like GPT-4 for evaluation is computationally expensive due to the extensive token usage required by elaborate prompts. For instance, the original GEMBA-MQM prompt typically requires between 1100 and 1200 tokens per example, accounting for exhaustive instructions and three few-shot examples. If we aim to evaluate this metric on 60k examples from the WMT22 Metrics Challenge Test Set~\citep{freitag-etal-2022-results}, the total token usage would be on the order of: 
\begin{multline*}
\text{Total Tokens} \approx 60\text{k} \times 1200 \approx 72\text{M} \text{ tokens.}
\end{multline*}

Given that GPT-4 pricing is \$10 per million tokens used, the estimated cost for evaluating the entire dataset would be:

\begin{multline*}
\text{Cost per full run} \approx \frac{72\text{M}}{1\text{M}} \times 10 \approx \$720.
\end{multline*}

These substantial costs present a barrier to practical deployment, especially in budget-constrained settings. It also makes LLM-based metrics like GEMBA-MQM egregiously expensive for large-scale evaluation scenarios, like online re-ranking of MT-systems outputs or web-scale dataset processing~\citep{peter-etal-2023-theres}.

To address these challenges, we introduce \textbf{\ourmodel{}},\footnote{\ourmodel{} stands for ``\textbf{Prompt} \textbf{Opt}imization for \textbf{Me}trics'' and is inspired by PrExMe \citep{Leiter2024PrExMeLS}, a recent method to investigate prompt exploration for MT evaluation metrics.} a novel approach to prompt optimization that reduces token usage and computational costs in LLM-based evaluation by utilizing a smaller, fine-tuned language model to compress input data without sacrificing necessary information. We summarize our contributions as follows:
{\bf a)} We propose a two-stage fine-tuning process for \textbf{\ourmodel{}}, involving supervised fine-tuning and preference optimization~\citep{hong2024reference}, to refine the model's outputs based on actual metric behavior with compressed inputs.
{\bf b)} We apply \textbf{\ourmodel{}} to MT, 
achieving a $2.32\times$ reduction in token usage for MT evaluation, without any loss in evaluation quality. 

Our work enhances the accessibility of advanced LLM-based evaluation metrics by making them more cost-effective and efficient for broader use in real-world NLG applications, thereby promoting diversity and inclusion in NLP research and application by enabling participation from under-resourced communities \citep{belouadi-eger-2023-uscore}.

\section{Related Work}
\label{sec:related_work}

\par The field of prompt optimization has seen substantial activity in recent years. This section covers works related to prompt optimization, efficient evaluation, and explainable metrics, highlighting how our approach builds upon and differs from existing research.

\par We begin by examining the landscape of \textbf{explainable evaluation metrics} \citep{kaster-etal-2021-global,opitz-frank-2021-towards,leiter-etal-2023-eval4nlp}. \citet{leiter2024towards} propose a taxonomy for such metrics, within which the GEMBA-MQM metric falls into the category of fine-grained error metrics. Naturally, these metrics critically depends on preservation of error-spans within source text and translation. Unlike xCOMET~\citep{guerreiro2023xcomet}, GEMBA-MQM's sentence-level scores are derived solely from extracted errors and their severities, adhering to the MQM guidelines~\citep{freitag-etal-2021-experts}. Due to this fact, we designed our method to fine-tune the model to preserve the error-spans in the source and translation text by utilizing the WMT MQM-annotated dataset. Further, \ourmodel{} can also be seen as explaining prompt-based evaluation metrics because it yields additional insights into which prompt parts are relevant for a metric to perform with high-quality.

\par In the realm of \textbf{prompt compression and optimization}, several notable approaches have been developed. LLMLingua~\citep{jiang-etal-2023-llmlingua} introduces a coarse-to-fine prompt compression method, employing a budget controller and an iterative token-level compression algorithm. This approach utilizes a smaller language model to compute token importance, substantially reducing token usage and inference latency across various tasks. Building upon this foundation, LLMLingua-2~\citep{pan-etal-2024-llmlingua} proposes a data distillation procedure to derive knowledge from LLMs for efficient and faithful task-agnostic prompt compression. By formulating prompt compression as a token classification problem and leveraging a Transformer encoder, LLMLingua-2 further improves compression efficiency and generalizability. Unlike both of those approaches, we specifically target the task of LLM-based evaluation metrics. We incorporate preference optimization based on actual metric behavior, tailoring the prompt compression to preserve essential evaluation information.

\par Black-Box Prompt Optimization (BPO)~\citep{cheng-etal-2024-black} applies prompt optimization for black-box LLM alignment, using a prompt preference optimizer trained on human preference data. This model-agnostic approach enhances the alignment of LLM outputs with human intents without requiring access to model parameters or additional training. Our method shares similarities in leveraging preference learning, but we extract these preferences by exploiting measurable quality differences during prompt execution with optimized outputs.

\par PRewrite~\citep{kong-etal-2024-prewrite} proposes an automated prompt engineering approach using reinforcement learning to rewrite prompts for improved prediction quality. While both PRewrite and our method optimize prompts without modifying the underlying language model, our approach distinguishes itself by simultaneously striving to reduce prompt size and improve quality.

\par  In the domain of \textbf{efficient MT evaluation metrics}, several approaches have been proposed to address the computational challenges posed by large trained metrics. xCOMET-lite~\citep{larionov2024xcomet}, COMETinho~\citep{rei-etal-2022-searching}, and FrugalScore~\citep{kamal-eddine-etal-2022-frugalscore} all employ techniques such as pruning, quantization, and distillation to create smaller, more efficient models. While these methods achieve impressive compression ratios, resulting in models with under 500M parameters, they often experience a drop in correlation with human judgment. Our approach aims to mitigate this quality loss by continuing to rely on large backbone LLMs while improving efficiency through targeted prompt optimization.

\par \citet{larionov-etal-2023-effeval} explore an alternative approach to creating an efficient version of BERTScore and MoverScore by replacing the underlying encoder model with smaller alternatives, such as pruned or distilled models. We touch this subject too, by evaluating our prompt compression model not only on large backbone LLM GPT-4o, but also on smaller models: GPT-4o mini and LLaMa 3.2. As we demonstrate in Section~\ref{sec:results}, our approach proves effective across different model sizes, improving evaluation efficiency even on smaller LLMs.
\section{Method}
\label{sec:method}

In this section, we detail our two-stage approach for prompt optimization aimed at reducing token usage and computational costs in LLM-based evaluation metrics. We first look into the optimization of prompt inputs (source texts and machine translations) and later detail our approach to compressing the rest of the prompt. The first stage involves supervised fine-tuning of a language model to learn the specifics of the prompt compression task. The second stage employs preference optimization using the Odds-Ratio Preference Optimization (ORPO) algorithm~\citep{hong2024reference} to refine the model's outputs based on the observed behavior of the metric. We chose this algorithm due to its specific focus on reducing the likelihood of generating the rejected responses, which are, in our case, low-quality compressed source and translation texts. The overall approach is visualized in Figure~\ref{fig:method}.

\subsection{Stage One: Supervised Fine-Tuning} 

In the first stage, we perform supervised fine-tuning of a language model to enable it to effectively compress input texts while preserving essential information required for accurate evaluation. In the context of the MT evaluation task, the model is trained to accept original uncompressed source texts and their respective machine translations and to generate three outputs:

\begin{enumerate}
    \item A compression rate $r$, selected as a floating-point number from the set $\mathcal{R}_{comp} = \{0.3, 0.4, 0.5, 0.6, 0.7, 0.8, 0.9, 1.0\}$, where $1.0$ indicates no compression.
    \item A list of potential substrings from the source and MT texts that contain translation errors.
    \item The compressed versions of the source and MT texts.
\end{enumerate}

We achieve two primary objectives by prompting the model to extract potential error spans from the source and MT texts. First, this process teaches the model to actively search for translation errors, thereby enhancing its understanding of MT evaluation. Second, by conditioning the compression of the texts on these identified error spans, the model ensures that critical information related to translation errors is preserved in the compressed outputs. This approach allows the compressed texts to retain essential details necessary for accurate evaluation, even with reduced token usage.

To construct the training dataset for supervised fine-tuning, we utilize Multidimensional Quality Metrics (MQM)-annotated data from the WMT Metrics shared tasks~\citep{freitag-etal-2021-results}. For each example in the dataset, we extract error-annotated spans based on the MQM annotations. We then perform random token removal to generate compressed texts, ensuring that the identified error spans remain intact in the compressed versions. The compression rate $r$ is randomly selected from the predefined set for each example.

Formally, let $S$ denote the original source text, $T$ the original MT text, and $E$ the set of error spans extracted from MQM annotations. The compressed source text $S'$ and compressed MT text $T'$ are generated by removing tokens from $S$ and $T$, respectively, while preserving all tokens within $E$.

This supervised fine-tuning process trains the model to perform the prompt compression task effectively, balancing the reduction in token count with the retention of essential information required for accurate evaluation.

\subsection{Stage Two: Preference Optimization}

In the second stage, we apply ORPO to further tune the model based on preferences between various compressions, enabling it to select the optimal compression for each example. The goal is to optimize the trade-off between token reduction and the preservation of evaluation quality.

To create the preference dataset needed for ORPO, we generate compressed versions of each example using the previously defined set of compression rates. For each compression rate $r$, we obtain compressed texts $S'_r$ and $T'_r$ using the fine-tuned model from Stage One. We then incorporate these compressed texts into the original GEMBA-MQM prompt and submit them to GPT-4o to obtain evaluation scores $s_r$.

The evaluation scores are compared to the score $s_{1.0}$ obtained from the uncompressed texts (with $r = 1.0$). For each example, we identify:

\begin{itemize}
    \item The \emph{chosen} compression with rate $r_{chosen}$ whose evaluation score $s_{r_{chosen}}$ has the minimal absolute difference from $s_{1.0}$. If multiple compression rates have equally minimal differences, we select the one with the lowest $r$ to prioritize higher compression.
    \item The \emph{rejected} compression with rate $r_{rejected}$, whose evaluation score $s_{r_{rejected}}$ has the maximal absolute difference from $s_{1.0}$. If multiple compression rates have equally maximal differences, we select the one with highest $r$.
\end{itemize}

Formally, we define the absolute score difference $\Delta_r = |s_r - s_{1.0}|$.
The chosen and rejected compression rates satisfy:
\begin{align*} 
    r_{chosen} & = \arg\min_{r \in \mathcal{R}_{comp}}\Delta_r, \\
    r_{rejected} & = \arg\max_{r \in \mathcal{R}_{comp}}\Delta_r.
\end{align*}

Using these preferences, we train the model from Stage One using the ORPO algorithm, which adjusts the model's parameters to increase the likelihood of generating outputs corresponding to the chosen compression rates over the rejected ones. The ORPO loss function is defined based on the odds ratio of the probabilities assigned to the chosen and rejected outputs:

\begin{align*}
\mathcal{L}_{\text{ORPO}} & = \mathbb{E}_{(x, y_c, y_r)}\left(\mathcal{L}_{\text{SFT}} + \lambda\cdot\mathcal{L}_{\text{OR}}\right) \\
\mathcal{L}_{\text{OR}} & = -\log\sigma\left(\log\frac{odds_\theta(y_c|x)}{odds_\theta(y_r|x)}\right) \\
odds_\theta(y|x) & = \frac{P_\theta(y|x)}{1-P_\theta(y|x)}
\end{align*}

where $x, y$ are a prompt and a completion respectively, $P_\theta(y_c|x)$ and $P_\theta(y_r|x)$ are the probabilities assigned by the model with parameters $\theta$ to the chosen and rejected outputs, respectively. $\mathcal{L}_{SFT}$ refers to standard cross-entropy loss used for language modeling.

This preference optimization process teaches the model to select compressions that yield evaluation scores closest to those obtained with uncompressed texts, effectively maintaining evaluation quality while reducing token usage. By prioritizing compression rates that minimize score discrepancies, the model learns to balance the trade-off between efficiency and quality.

\subsection{Simplified Prompts}
\label{sec:method/simplified-prompts}

In the context of MT evaluation, the source texts and their translations are typically short, often comprising up to two or three sentences. However, the GEMBA-MQM prompt includes long instructions, outlining the entire MQM error typology. Those instructions are included in both few-shot examples and in the target example. We hypothesize that this surplus of tokens raises computational costs without necessarily enhancing evaluation quality.

Initially, we also trained a prompt optimization model using preference data to compress the instruction component of the prompt. However, upon evaluation, we found that the compressed instructions generated were effectively the same across different inputs. This uniformity suggested that a fixed simplified instruction could suffice without dynamic compression per example.

Moreover, as demonstrated in Section~\ref{sec:experiments}, the use of these simplified instructions did not adversely affect the metric quality. The evaluation quality remained comparable to that achieved with the original, more verbose GEMBA-MQM prompt.

Consequently, we adopt a fixed simplified instruction template for our evaluations and discontinued further instruction compression. For reference, both the original GEMBA-MQM prompt and our simplified version are provided in the Appendix~\ref{sec:appendix/gemba-prompt}.
\section{Experiments}
\label{sec:experiments}

In this section, we evaluate the effectiveness of our proposed prompt optimization approach in reducing token usage while maintaining evaluation quality in LLM-based evaluation metrics. We conduct experiments on MT evaluation and assess the compression efficiency and the impact on evaluation quality.

As a base model for fine-tuning, we utilize the LLaMA-3.2~\citep{dubey2024llama} model in two variants --- 1B and 3B parameters --- chosen for its extensive vocabulary and multilingual pre-training beneficial for handling diverse language pairs. 

\paragraph{Stage One: Supervised Fine-Tuning}

For the first stage, we use the WMT Metrics shared task datasets with MQM annotations from the years 2020 to 2022~\cite{freitag-etal-2021-results}. From this corpus, we exclude a test subset of approximately 16k examples from the news domain from the year 2022 for evaluation purposes. The remaining training data consists of approximately 145k examples across three language pairs: English-Russian, English-German, and Chinese-English. 

Following the approach described in Section~\ref{sec:method}, we construct prompts with system instructions for the model. The prompt follows this template:

\begin{quote}
\texttt{System: You are a helpful AI assistant that intelligently compresses and summarizes the given Machine Translation outputs for further evaluation.}

\texttt{User: Compress the following MT input and output:}
\texttt{Source:```\{\textit{src-text}\}```}
\texttt{MT:```\{\textit{mt-text}\}```}
\end{quote}

The expected model completion is:

\begin{quote}
\texttt{Assistant:}\texttt{Rate = } $r$

\texttt{Quality-relevant parts of Source:} [\textit{error-span-1}]; ... ;[\textit{error-span-N}] \texttt{or None}

\texttt{Quality-relevant parts of Translation:} [\textit{error-span-1}]; ...;  [\textit{error-span-N}] \texttt{or None}

\texttt{Compressed Source:}```\{\textit{compressed-src}\}```

\texttt{Compressed MT:}```\{\textit{compressed-mt}\}```
\end{quote}


Here, \{\textit{src-text}\} and \{\textit{mt-text}\} denote the original source and machine-translated texts. We provide an example of the text compression with trained model in Figure~\ref{lst:compression}, where we demonstrate both model inputs --- source text and machine translation --- as well as the outputs --- compression rate, error spans and compressed texts.

\begin{figure}
\small{
\textcolor{SeaGreen}{<model-prompt>}\\
\textcolor{blue}{Example Source}: 
So they're crossing thousands of kilometers in a fraction of a second, and as they do so, they not only curve space, but they leave behind in their wake a ringing of space, an actual wave on space-time.\\
\textcolor{red}{Example MT}: Sie überqueren also in Sekundenbruchteilen Tausende von Kilometern und krümmen dabei nicht nur den Raum, sondern hinterlassen auch ein Klingeln des Raums, eine tatsächliche Welle der Raumzeit.\\
\textcolor{SeaGreen}{</model-prompt>}\\
\textcolor{BurntOrange}{Example Reference Translation}: Sie können Tausende von Kilometern im Bruchteil einer Sekunde zurücklegen. Und während sie das tun, beugen sie nicht nur den Raum, sondern sie hinterlassen in ihrem Kielwasser ein Klingen des Raumes, eine echte Welle aus Raumzeit.\\
\textcolor{Mulberry}{Example Gold MQM Annotations}: [{'start': 4, 'end': 14, 'severity': 'minor', 'text': 'überqueren'}, {'start': 178, 'end': 179, 'severity': 'minor', 'text': ' '}]

\textcolor{SeaGreen}{<model-generation>}\\
\textcolor{Bittersweet}{Rate}: 0.5

\textcolor{Mulberry}{Quality-relevant parts of Source}: None

\textcolor{Mulberry}{Quality-relevant parts of MT}: [Sie überqueren also in Sekundenbruchteilen Tausende von Kilometern];

\textcolor{blue}{Compressed Source}: ```thousands kilometers fraction of a second curve spaceringing of spacean actual wave```

\textcolor{red}{Compressed MT}: ```Sie überqueren Sekundenbruchteilen Tausende von Kilometern krümmen nicht eine tatsächliche Welle```\\
\textcolor{SeaGreen}{</model-generation>}
}
\caption{Example of Model-generated compression. Reference translation and Gold MQM annotations are provided for readers convenience and are not part of the model prompt.}
\label{lst:compression}
\end{figure}

We train the model for one epoch using a batch size of 64 and a learning rate of $2 \times 10^{-5}$ with cosine warmup over 6\% of the training steps. We apply a weight decay of 0.01. For efficient training, we use Low-Rank Adaptation (LoRA) adapters~\citep{DBLP:conf/iclr/HuSWALWWC22}, with the following parameters: rank $r = 32$, scaling factor $\alpha = 16$, and dropout rate of 0.5.

\paragraph{Stage Two: Preference Optimization}

For the second stage, we construct a preference dataset by selecting a subset of 20\se{k} examples from the training data. For each example, we generate compressed versions at each compression rate from our predefined set, resulting in eight different compressions per example.

We incorporate these compressed examples into the original GEMBA-MQM prompt and evaluate them using the GPT-4o model to obtain evaluation scores. Based on these scores, we select the \emph{chosen} and \emph{rejected} compressions following the procedure described in Section~\ref{sec:method}. 

Using this preference dataset, we fine-tune the model from Stage One employing ORPO. We train for three epochs with a batch size of 64, a learning rate of $1 \times 10^{-5}$, and cosine warmup over 6\% of training steps. We set the ORPO $\lambda$ parameter to 0.1 following the original paper authors.

\paragraph{Evaluation Procedure}

To evaluate the effectiveness of our model, we apply it to compress the examples in the test set. This compression is performed on both the few-shot examples and the target example within the GEMBA-MQM prompts. We then conduct MT evaluation using these compressed prompts with proprietary LLMs GPT-4o and GPT-4o-mini and the openly-available LLaMA 3.2 (90B-Instruct) version. In our experiments, we utilize both the original and simplified prompts (as described in Section~\ref{sec:method/simplified-prompts}) and generate outputs in both plain text format and in JSON.

We assess the quality of the evaluations by computing pairwise accuracy at the system level, as suggested by~\citet{deutsch-etal-2023-ties}, comparing our results with human judgments. Additionally, we calculate segment-level Kendall's~$\tau$ correlations for each language pair to measure the agreement between our evaluations and human assessments.

To quantify the efficiency gains, we measure the cumulative token usage during the evaluations. This provides insights into the computational cost savings achieved through our prompt optimization approach. To compare the effectiveness of our approach, we also include two baseline results obtained using the LLMLingua-2~\citep{pan-etal-2024-llmlingua} prompt compression method. In those cases, we have applied \textit{microsoft/llmlingua-2-xlm-roberta-large-meetingbank} to compress source texts and machine translations similarly as we do with PromptOptMe. We fix and test two compression rates: 30\% and 50\%.
\section{Results}
\label{sec:results}

In this section, we present the results of our experiments evaluating the effectiveness of our proposed prompt optimization approach.

\begin{table*}[htp]
\small
\centering
\begin{tabularx}{\textwidth}{lcccccc}
\toprule
\textbf{Model + Prompt} & \textbf{Token Usage} & \textbf{Reduction Rate} & \textbf{Pairwise Accuracy} & \textbf{En-Ru $\tau$} & \textbf{En-De $\tau$} & \textbf{Zh-En $\tau$} \\
\midrule
GPT-4o ref & 19M & 1.00 & 0.7789 & 0.4365 & 0.3950 & 0.3692 \\
GPT-4o mini ref & 19M & 1.00 & 0.7631 & 0.3723 & 0.3165 & 0.3472 \\
LLaMa3.2-90B ref & 20M & 1.00 & 0.7526 & 0.3416 & 0.2920 & 0.3576 \\
GPT-4o lite & 10.4M & 1.84 & 0.7736 & 0.3838 & 0.3207 & 0.2890 \\
\midrule
GPT-4o lite &  &  &  &  &  &  \\
\quad \ourmodel{}-3B & 8.07M & 2.37 & 0.7736 & \textbf{0.4455} & \textbf{0.4065} & \textbf{0.3738} \\
\quad \ourmodel{}-1B & 8.8M & 2.15 & 0.7644 & 0.4122 & 0.3900 & 0.3743 \\
\midrule
GPT-4o mini lite &  &  &  &  &  &  \\
\quad \ourmodel{}-3B & 8.07M & 2.37 & 0.7842 & 0.3177 & 0.3238 & 0.3596 \\
\quad \ourmodel{}-1B & 8.8M & 2.15 & 0.7531 & 0.3089 & 0.3125 & 0.3468 \\
\midrule
LLaMa3.2-90B lite &  &  &  &  &  &  \\
\quad \ourmodel{}-3B & 8.8M & 2.27 & 0.7526 & 0.3505 & 0.3191 & 0.3123 \\
\quad \ourmodel{}-1B & 9.0M & 2.22 & 0.7345 & 0.3203 & 0.2891 & 0.3000 \\
\midrule
GPT-4o lite &  &  &  &  &  &  \\
\quad LLMLingua2 @ 50\% & 7.6M & 2.5 & 0.4736 & 0.0055 & 0.0492 & 0.1247 \\
\quad LLMLingua2 @ 30\% & 7.0M & 2.7 & 0.5421 & 0.00 & 0.03 & 0.1949 \\
\bottomrule
\end{tabularx}
\caption{Evaluation results for machine translation with different prompting strategies and models. \textbf{GPT-4o ref} refers to the original GPT-4 with the full (uncompressed) GEMBA-MQM prompt. \textbf{lite} denotes the simplified prompting approach with JSON-formatted output. \ourmodel{}-3B and \ourmodel{}-1B represent our prompt optimization models based on LLaMA 3.2 with 3B and 1B parameters, respectively, used for input compression. \textbf{Token Usage} shows the total token usage. \textbf{Reduction Rate} is the token reduction rate compared to the baseline (\textbf{-ref} for each model). \textbf{Pairwise Accuracy} stands for pairwise system-level accuracy. \textbf{En-Ru $\tau$}, \textbf{En-De $\tau$}, and \textbf{Zh-En $\tau$} refer to the segment-level Kendall~$\tau$ correlations for English-Russian, English-German, and Chinese-English language pairs, respectively. Boldface indicates the best quality in each column.}
\label{tab:mt_results}
\end{table*}

As shown in Table~\ref{tab:mt_results}, the baseline model, \textbf{GPT-4o ref}, uses GPT-4o with the full (uncompressed) GEMBA-MQM prompt, ends up with a total amount of 19M input tokens used for entire test set of 16k examples and serving as a reference point. The simplified prompting approach from Section~\ref{sec:method/simplified-prompts}, denoted as \textbf{GPT-4o lite}, reduces the token usage to 10.4M tokens, achieving a reduction rate of 2.04$\times$ while maintaining a similar pairwise accuracy of 0.7736, but lower segment-level Kendall~$\tau$ correlations across the language pairs (0.3838 for En-Ru, 0.3207 for En-De, and 0.2890 for Zh-En).

By applying our prompt optimization model, \ourmodel{}-3B, with \textbf{GPT-4o lite} prompt template, we achieve the highest reduction rate of 2.37$\times$, reducing token usage from 19M to 8.3M tokens. Surprisingly, \ourmodel{}-3B attains the best segment-level Kendall~$\tau$ correlations across all language pairs, with 0.4455 for En-Ru, 0.4065 for En-De, and 0.3738 for Zh-En, fully recovering reduced scores in the baseline, while also retaining the same level of system-level pairwise accuracy.

This trend continues with other backbone LLMs. For instance, using \ourmodel{}-3B with \textbf{GPT-4o mini lite} prompt, we, again, achieve the same reduction rate of $2.37\times$ and observe an improvement in pairwise accuracy to \textbf{0.7842}, which is a 2.77\% increase over the \textbf{GPT-4o mini ref} baseline. However, the segment-level correlations exhibit mixed results: while there is a substantial decrease for En-Ru (14.67\% lower than the baseline), there is a positive improvement for En-De (an increase of 2.30\%) and Zh-En (a 3.57\% increase).

Similarly, for the \textbf{LLaMa3.2-90B} model, applying \ourmodel{}-3B results in a reduction rate of $2.27\times$. The pairwise accuracy remains the same as the baseline (0.7526), but the segment-level Kendall~$\tau$ correlations show modest improvements for En-Ru (a 2.61\% increase) and En-De (9.28\% higher), while Zh-En experiences a decrease of 12.68\%.

Thus, overall, we see substantial efficiency gains and simultaneously often an increase in quality of the resulting metrics; however, there are cases when metric quality reduces, sometimes considerably.

In comparison with the baseline approaches with the LLMLingua-2, we see dramatically decreased quality in both compression settings. For the system-level pairwise accuracy, we observe a 31\%-40\% decrease from the original uncompressed metric. On a segment level, the quality decrease is even more catastrophic. For two out of three language pairs, the correlation plunges to near zero. For Zh-En, it drops to 0.19-0.12. This indicates that LLMLingua-2 catastrophically damages the prompt to the point of complete non-usability.

\section{Discussion}
\label{sec:discussion}

Our proposed prompt optimization model, \ourmodel{}, demonstrates substantial improvements in computational efficiency for machine translation evaluation. By integrating \ourmodel{} with a simplified prompting approach, we achieve substantial reductions in token usage—up to 2.37$\times$ less than the baseline—while maintaining or even improving evaluation quality across various metrics. These findings suggest that it is possible to compress inputs substantially while maintaining evaluation quality. We speculate that outputs of few-shot examples, which are not compressed in any way in our experiments, play a crucial role in establishing model quality in MT evaluation, as they themselves cover all error severity levels as well as error categories from MQM typology. Preserving entire source and translation texts, in turn, appears to be redundant. We are able to retain quality while preserving only part of the text, which includes error-spans.

The superior quality of \ourmodel{}-3B, compared to the smaller \ourmodel{}-1B model, 
indicates that a larger prompt optimization model is more effective at compressing input data efficiently. Specifically, \ourmodel{}-3B refers to our prompt optimization model based on LLaMA~3.2 with 3~billion parameters, achieves 10\% (2.37 vs. 2.15) higher compression rate on average, along with 2\%-3\% higher pairwise accuracy, compared to \ourmodel{}-1B. It demonstrates the importance of model capacity in the compression process. Further experiments with larger LLMs could potentially help get a better understanding of scaling laws for that particular task.

When applying \ourmodel{}-3B to different backbone LLMs, such as \textbf{GPT-4o mini} and \textbf{LLaMa3.2-90B}, we observe varying results. For \textbf{GPT-4o mini}, we achieve an improved pairwise accuracy compared to the baseline, with a reduction rate of 2.37$\times$. In contrast, when applying \ourmodel{}-3B to \textbf{LLaMa3.2-90B}, the pairwise accuracy remains consistent with the baseline, and we observe modest improvements in segment-level Kendall~$\tau$ correlations for some language pairs. In both cases, we notice a slight degradation in segment-level correlations in one of the language pairs, however. Nonetheless, we can conclude that our prompt optimization model achieves generalization across different backbone LLMs, despite being trained only on preferences obtained through GPT-4o.

Additionally, compared to baseline approaches using LLMLingua-2, our method clearly outperforms in both system-level and segment-level evaluations. The substantial quality degradation observed with LLMLingua-2, particularly at the segment level, indicates that it may not be suitable for effective prompt compression in MT evaluation tasks. This result is in agreement with our hypothesis that preserving quality-relevant spans in text is essential for maintaining evaluation quality. LLMLingua-2, as a non-task-specific prompt compression method, is not trained to take that into account.

\section{Conclusion}
\label{sec:conclusion}

In this paper, we introduced \textbf{\ourmodel{}}, a prompt optimization approach designed to reduce token usage and computational costs in large language model-based evaluation metrics. By leveraging a smaller, fine-tuned language model to compress input data, we achieved substantial reductions in token usage without compromising evaluation quality. Specifically, \textbf{\ourmodel{}} reduced token usage by up to $2.37\times$ while maintaining or improving segment-level Kendall~$\tau$ correlations across multiple language pairs as well as system-level pairwise accuracy.

Our approach enhances the practicality of LLM-based evaluation metrics, making them more cost-effective and accessible for large-scale natural language generation (NLG) applications. By addressing the computational expense associated with extensive prompt token usage, \textbf{\ourmodel{}} enables more efficient evaluations. This reduction in computational cost makes state-of-the-art MT evaluation more accessible to under-resourced researchers and students, who may have limited access to computational resources or funding for extensive LLM usage. By lowering the barriers to high-quality evaluation, \textbf{\ourmodel{}} democratizes the ability to conduct advanced research and development in machine translation. Our approach is especially useful in this case, as it does not require a compromise in metric quality, offering no quality drop while decreasing the token usage. Moreover, we are able to effectively compress the inputs for smaller LLMs as well, such as GPT-4o mini and LLaMa 3.2 90B, which makes SOTA evaluation metrics even more accessible.

While our experiments focused on machine translation, where error highlights and annotations are common, we acknowledge that extending this approach to other NLG tasks may present challenges, particularly in scenarios where such detailed error annotations are not available. Future work could explore adapting \textbf{\ourmodel{}} to other NLG evaluation tasks, investigating how input compression affects evaluation quality in contexts without explicit error spans, and determining the generalizability of our method across various domains.

It is also important to note, that the optimized prompts generated by \textbf{\ourmodel{}} are not guaranteed to strictly adhere to the MQM error typology. This divergence from the standard evaluation framework may impact the consistency and interpretability of the evaluation results, especially when comparing outputs across different systems or studies. Future work should focus on enhancing the alignment of compressed prompts with established evaluation standards to improve the reliability and comparability of the assessments.

We plan to release the code and models to the public at \url{https://github.com/NL2G/promptoptme} to facilitate further research and application in this area. We believe that \textbf{\ourmodel{}} offers a promising direction for efficient high-quality LLM-based evaluations.

\section*{Limitations}
\label{sec:limitations}

While our proposed approach demonstrates substantial reductions in token usage without sacrificing evaluation quality, there are several limitations to this study that we acknowledge.

First, our experiments are primarily focused on machine translation MT evaluation. Although we achieved notable efficiency gains in this domain, we have not extensively tested the applicability of \textbf{\ourmodel{}} on other NLG tasks. Further research is necessary to validate the effectiveness of \textbf{\ourmodel{}} across a broader spectrum of NLG applications.

Second, we evaluated our method using only one openly available language model in addition to \textbf{GPT-4o}, specifically the \textbf{LLaMA-3.2} model. While the results are promising, testing our approach on a wider variety of models, including more open-source and proprietary LLMs with diverse architectures and sizes, would help establish the generalizability and robustness of \textbf{\ourmodel{}}. Future work should include experiments with additional models to better understand the applicability of our prompt optimization technique.

Additionally, our approach relies on a two-stage fine-tuning process involving supervised fine-tuning and preference optimization. This process requires access to sufficient training data and computational resources for fine-tuning the smaller language model used for prompt compression. In scenarios where such resources are limited or unavailable, the practicality of our method may be constrained. Exploring alternative approaches that require less extensive fine-tuning or that leverage zero-shot or few-shot learning could mitigate this limitation. In addition to one-time training cost, practitioners using \ourmodel{} should also consider inference costs for the compression model itself. Further work should take these costs into account for more fair comparison with the baseline.

Furthermore, while our prompt optimization aims to preserve essential evaluation information, there is a possibility that some fine-grained details, such as specific MQM error categories, may not be fully captured in the compressed prompts. This could potentially affect the precision in identifying certain types of translation errors, leading to less detailed evaluations. Ensuring that critical information is retained during compression without increasing token usage remains a challenge that warrants further investigation.

\bibliography{anthology,custom}

\appendix

\section{Original and Simplified GEMBA-MQM prompt}
\label{sec:appendix/gemba-prompt}
\begin{figure*}[tp]
\tiny{
\textcolor{Periwinkle}{================================ System Message =================================}\\
You are an annotator for the quality of machine translation. Your task is to identify errors and assess the quality of the translation.\\
\textcolor{Emerald}{================================ Human Message =================================}\\
English source:\\
```I do apologise about this, we must gain permission from the account holder to discuss an order with another person, I apologise if this was done previously, however, I would not be able to discuss this with yourself without the account holders permission.```\\
German translation:\\
```Ich entschuldige mich dafür, wir müssen die Erlaubnis einholen, um eine Bestellung mit einer anderen Person zu besprechen. Ich entschuldige mich, falls dies zuvor geschehen wäre, aber ohne die Erlaubnis des Kontoinhabers wäre ich nicht in der Lage, dies mit dir involvement.```\\

Based on the source segment and machine translation surrounded with triple backticks, identify error types in the translation and classify them. The categories of errors are: accuracy (addition, mistranslation, omission, untranslated text), fluency (character encoding, grammar, inconsistency, punctuation, register, spelling), style (awkward), terminology (inappropriate for context, inconsistent use), non-translation, other, or no-error.
Each error is classified as one of three categories: critical, major, and minor. Critical errors inhibit comprehension of the text. Major errors disrupt the flow, but what the text is trying to say is still understandable. Minor errors are technically errors, but do not disrupt the flow or hinder comprehension.\\
\textcolor{RedViolet}{================================== AI Message ==================================}\\
Critical:\\
no-error\\
Major:\\
accuracy/mistranslation - "involvement"\\
accuracy/omission - "the account holder"\\
Minor:\\
fluency/grammar - "wäre"\\
fluency/register - "dir"\\
\textcolor{Emerald}{================================ Human Message =================================}\\
English source:\\
```Talks have resumed in Vienna to try to revive the nuclear pact, with both sides trying to gauge the prospects of success after the latest exchanges in the stop-start negotiations.```\\
Czech translation:\\
```Ve Vídni se ve Vídni obnovily rozhovory o oživení jaderného paktu, přičemž obě partaje se snaží posoudit vyhlídky na úspěch po posledních výměnách v jednáních.```\\

Based on the source segment and machine translation surrounded with triple backticks, identify error types in the translation and classify them. The categories of errors are: accuracy (addition, mistranslation, omission, untranslated text), fluency (character encoding, grammar, inconsistency, punctuation, register, spelling), style (awkward), terminology (inappropriate for context, inconsistent use), non-translation, other, or no-error.
Each error is classified as one of three categories: critical, major, and minor. Critical errors inhibit comprehension of the text. Major errors disrupt the flow, but what the text is trying to say is still understandable. Minor errors are technically errors, but do not disrupt the flow or hinder comprehension.\\
\textcolor{RedViolet}{================================== AI Message ==================================}\\
Critical:\\
no-error\\
Major:\\
accuracy/addition - "ve Vídni"\\
accuracy/omission - "the stop-start"\\
Minor:\\
terminology/inappropriate for context - "partaje"\\
\textcolor{Emerald}{================================ Human Message =================================}\\
Chinese source:\\
```\begin{CJK*}{UTF8}{gbsn}大众点评乌鲁木齐家居卖场频道为您提供高铁居然之家地址，电话，营业时间等最新商户信息，找装修公司，就上大众点评\end{CJK*}```\\
English translation:\\
```Urumqi Home Furnishing Store Channel provides you with the latest business information such as the address, telephone number, business hours, etc., of high-speed rail, and find a decoration company, and go to the reviews.```\\

Based on the source segment and machine translation surrounded with triple backticks, identify error types in the translation and classify them. The categories of errors are: accuracy (addition, mistranslation, omission, untranslated text), fluency (character encoding, grammar, inconsistency, punctuation, register, spelling), style (awkward), terminology (inappropriate for context, inconsistent use), non-translation, other, or no-error.
Each error is classified as one of three categories: critical, major, and minor. Critical errors inhibit comprehension of the text. Major errors disrupt the flow, but what the text is trying to say is still understandable. Minor errors are technically errors, but do not disrupt the flow or hinder comprehension.\\
\textcolor{RedViolet}{================================== AI Message ==================================}\\
Critical:\\
accuracy/addition - "of high-speed rail"\\
Major:\\
accuracy/mistranslation - "go to the reviews"\\
Minor:\\
style/awkward - "etc.,"\\
\textcolor{Emerald}{================================ Human Message =================================}\\
\{source\_lang\} source:\\
```\{source\_seg\}```\\
\{target\_lang\} translation:\\
```\{target\_seg\}```\\

Based on the source segment and machine translation surrounded with triple backticks, identify error types in the translation and classify them. The categories of errors are: accuracy (addition, mistranslation, omission, untranslated text), fluency (character encoding, grammar, inconsistency, punctuation, register, spelling), style (awkward), terminology (inappropriate for context, inconsistent use), non-translation, other, or no-error.
Each error is classified as one of three categories: critical, major, and minor. Critical errors inhibit comprehension of the text. Major errors disrupt the flow, but what the text is trying to say is still understandable. Minor errors are technically errors, but do not disrupt the flow or hinder comprehension.
}
\caption{Original Prompt for GEMBA-MQM metric.}
\label{lst:gemba-original}
\end{figure*}

\begin{figure*}[tp]
\tiny{
\textcolor{Periwinkle}{================================ System Message =================================}\\
Identify and categorize translation errors. Respond in JSON.\\
\textcolor{Emerald}{================================ Human Message =================================}\\
English: ```I do apologise about this, we must gain permission from the account holder to discuss an order with another person, I apologise if this was done previously, however, I would not be able to discuss this with yourself without the account holders permission.```\\
German: ```Ich entschuldige mich dafür, wir müssen die Erlaubnis einholen, um eine Bestellung mit einer anderen Person zu besprechen. Ich entschuldige mich, falls dies zuvor geschehen wäre, aber ohne die Erlaubnis des Kontoinhabers wäre ich nicht in der Lage, dies mit dir involvement.```;\\
Errors?\\
\textcolor{RedViolet}{================================== AI Message ==================================}\\
{"critical": ["no-error"], "major": [{"accuracy/mistranslation": "involvement"}, {"accuracy/omission": "the account holder"}], "minor": [{"fluency/grammar": "wäre"}, {"fluency/register": "dir"}]}\\
\textcolor{Emerald}{================================ Human Message =================================}\\
English: ```Talks have resumed in Vienna to try to revive the nuclear pact, with both sides trying to gauge the prospects of success after the latest exchanges in the stop-start negotiations.```\\
Czech: ```Ve Vídni se ve Vídni obnovily rozhovory o oživení jaderného paktu, přičemž obě partaje se snaží posoudit 
vyhlídky na úspěch po posledních výměnách v jednáních.```;\\
Errors?\\
\textcolor{RedViolet}{================================== AI Message ==================================}\\
{"critical": ["no-error"], "major": [{"accuracy/addition": "ve Vídni"}, {"accuracy/omission": "the stop-start"}], "minor": [{"terminology/inappropriate for context": "partaje"}]}\\
\textcolor{Emerald}{================================ Human Message =================================}\\
Chinese: ```\begin{CJK*}{UTF8}{gbsn}大众点评乌鲁木齐家居卖场频道为您提供高铁居然之家地址，电话，营业时间等最新商户信息，找装修公司，就上大众点评\end{CJK*}```\\
English: ```Urumqi Home Furnishing Store Channel provides you with the latest business information such as the address, telephone number, business hours, etc., of high-speed rail, and find a decoration company, and go to the reviews.```;\\
Errors?\\
\textcolor{RedViolet}{================================== AI Message ==================================}\\
{"critical": [{"accuracy/addition": "of high-speed rail"}], "major": [{"accuracy/mistranslation": "go to the 
reviews"}], "minor": [{"style/awkward": "etc.,"}]}\\
\textcolor{Emerald}{================================ Human Message =================================}\\
\{source\_lang\}: ```\{source\_seg\}```\\
\{target\_lang\}: ```\{target\_seg\}```\\
Errors?\\
}
\caption{Simplifield Prompt for GEMBA-MQM metric.}
\label{lst:gemba-simplified}
\end{figure*}

\todo[disable]{SE: the 2nd one is not the original but the simplified, I assume. Then it has the wrong caption. Why does the original have three examples (German, Czech, Chinese?) and the simplified only one (English-German)?; DL: fixed}

\section{Additional Details}
\subsection{Package versions}
We use the following versions of key software packages in our environment:
\begin{itemize}
    \item Python OpenAI SDK: 1.42.0
    \item mt-metrics-eval: 0.0.3
    \item SciPy: 1.14.1
    \item LLMLingua: 0.2.2
\end{itemize}

\subsection{Artifact Intended Use and License}
To the best of our knoweledge, our use of scientific artifacts, namely Llama 3.2 pretrained model(s), is consistent with intended use policy and with the license. Those are outlined in the respective model card~\footnote{\url{https://huggingface.co/meta-llama/Llama-3.2-3B-Instruct}}.

\subsection{Total Computational Budget}

Our total computational budget for supervised finetuning and preference optimization of \ourmodel{} models is 186 GPU hours.

\end{document}